\pdfoutput=1
\documentclass[10pt,twocolumn,letterpaper]{article}

\usepackage{cvpr}
\usepackage{times}
\usepackage{epsfig}
\usepackage{graphicx}
\usepackage{amsmath}
\usepackage{amssymb}

\usepackage[pagebackref=true,breaklinks=true,letterpaper=true,colorlinks,bookmarks=false]{hyperref}

\usepackage[skip=0pt]{caption}
\usepackage{booktabs}
\usepackage{multirow}
\usepackage{ctable}

\usepackage{subcaption}
 \cvprfinalcopy 

\def\cvprPaperID{8} 

\ifcvprfinal\fi
\begin{document}

\newcommand\blfootnote[1]{%
  \begingroup
  \renewcommand\thefootnote{}\footnote{#1}%
  \addtocounter{footnote}{-1}%
  \endgroup
}

\title{Enhancing Salient Object Segmentation Through Attention}

\author{Anuj Pahuja* \quad Avishek Majumder* \quad Anirban Chakraborty \quad R. Venkatesh Babu\\
Indian Institute of Science, Bangalore, India\\
{\tt\small anujpahuja13@gmail.com \quad avishek.alex15@gmail.com \quad \{anirban,venky\}@iisc.ac.in}
}


\twocolumn[{%
\renewcommand\twocolumn[1][]{#1}%
\vspace{-1em}
\maketitle
\vspace{-1em}
\begin{center}
   \centering \includegraphics[width=\textwidth]{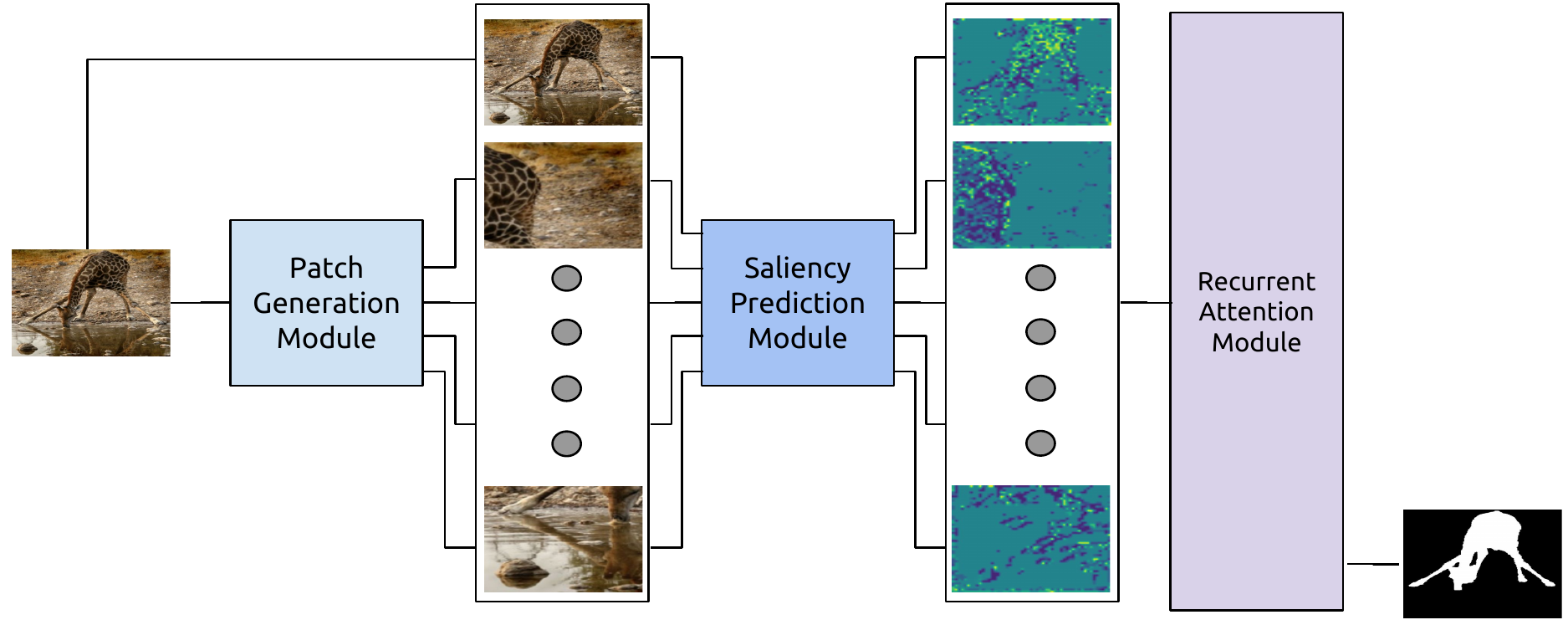} \captionof{figure}{Overview of the proposed approach. An input RGB image goes through three different modules. Patch Generation Module learns to generate image patches in a differentiable manner. The Saliency Prediction Module operates on every patch and generates saliency feature maps. Finally, the Recurrent Attention Module aggregates the bag of features and iteratively refines the complete segmentation map.}
   \label{fig:overview}
\end{center}%
}]

\blfootnote{*Equal contribution}

\begin{abstract}
   Segmenting salient objects in an image is an important vision task with ubiquitous applications. The problem becomes more challenging in the presence of a cluttered and textured background, low resolution and/or low contrast images. Even though existing algorithms perform well in segmenting most of the object(s) of interest, they often end up segmenting false positives due to resembling salient objects in the background. In this work, we tackle this problem by iteratively attending to image patches in a recurrent fashion and subsequently enhancing the predicted segmentation mask. Saliency features are estimated independently for every image patch which are further combined using an aggregation strategy based on a Convolutional Gated Recurrent Unit (ConvGRU) network. The proposed approach works in an end-to-end manner, removing background noise and false positives incrementally. Through extensive evaluation on various benchmark datasets, we show superior performance to the existing approaches without any post-processing.
\end{abstract}

\section{Introduction}

Saliency is an important aspect of human vision. It is the phenomenon that allows our brain to focus on some parts of a scene more than the rest of it. Thousands of years of evolution has optimized our brain usage by focusing only in the most important regions in our field of view and ignore the rest of it. Indeed, even in computer vision, saliency plays a huge role in many applications, including what humans use it for - compressed representation~\cite{guo2010novel,itti2004automatic}. Saliency can be exploited to improve agent navigation in the wild~\cite{craye2016environment}, image retrieval~\cite{cheng2017intelligent,gao20123,he2012mobile}, content based object re-targeting~\cite{cheng2010repfinder,setlur2005automatic}, scene parsing~\cite{zhao2015saliency}, object detection and segmentation~\cite{gao2005discriminant,liu2015predicting,pan2016shallow} among many others. Due to its vast applications in vision, saliency prediction is a well established problem with decades of on-going research. Despite the efforts, the problem still remains open due to factors like cluttered background, multiple instances of non-salient objects, scattered salient regions, low contrast scene, and the definition of saliency varying from application to application.

Salient object detection (SOD) or segmentation is an immediate extension of saliency prediction, as it requires a precise pixel-wise segmentation of the object of interest in the scene. This is a harder task than saliency prediction due to the amount of precision required. We observe that background is the primary reason for poor segmentations. Lack of a well-defined boundary between the salient object and the background can make it very difficult for vision algorithms to segregate objects accurately. Besides often being similar to the foreground object, part of the background can also contribute to saliency, which further affects the segmentation performance due to false predictions. Cluttered or texture-rich background is often the reason why saliency models may focus on the background, failing to segment out the true object of interest. All these challenges are inherently associated to the task of salient object segmentation from images in the wild.

In recent past, Convolutional Neural Networks (CNNs) have shown impressive performance on this task, achieving significant improvements over existing approaches, both in speed and accuracy. Although existing approaches succeed in segmenting out majority of the salient object(s), they often miss out on finer details and/or segment partly salient background regions during the global optimization process. We believe that individually attending to finer image regions or regions separating object and background can refine the overall segmentation mask. Since handcrafting such regions is a very subjective and unscalable approach, we propose to use a learnable module for estimating these region locations. We would also like to incorporate learned features from a region$_k$ for predicting the next region$_{k+1}$, whilst maintaining the spatial context and improving the whole saliency map. This symbiotic relationship can be best exploited using a recurrent network, where different regional features can act as a temporal sequence. Such a strategy could impart important foreground/background distinction to the network along with fine object details that can be aggregated and improved upon iteratively.    

We also take inspiration from recent approaches for the task of video object segmentation task. Motion patterns in a scene, specifically the differences between the foreground object motion and background motion may act as an important cue to segregate foreground from background. Moreover, the information flow within a temporal neighborhood often improves the segmentation accuracy especially in cases of occlusion and background clutter. Tokmakov et al.~\cite{LVO} leveraged such temporal dynamics in the video via feature aggregation using a gated recurrent network. Analogously, different regions/patches might emulate this behavior of spatio-temporal perturbations in a single image.

\textbf{Our Contributions:} In this work, we propose an end-to-end salient object detection architecture comprising of three modules that a) learns what image regions to attend, b) effectively aggregates the learned features from regions, and c) incrementally refines the overall segmentation in an interpretable way. Patch Generation Module (PGM) learns and crops the desired regions in a differentiable way, creating a bag of images (including the input image). Saliency Prediction Network (SPN) outputs the saliency features of each image in the bag independently. Recurrent Attention Module (RAM) combines these features using a novel aggregation strategy based on a pair of encoder-decoder Convolutional GRUs. Through our intuitive approach, we achieve state-of-the-art results on challenging SOD datasets.

\section{Related Work}

Given the ubiquity of the salient object segmentation problem, a lot of approaches have been tried and tested in the past literature. Earlier approaches mostly work on various low-level hand-crafted features in a data-independent setup. Object and background priors~\cite{shen2012unified}, global regional contrast~\cite{cheng2013efficient,jiang2013submodular,cheng2015global} and boundary priors~\cite{cheng2013efficient,zhu2014saliency} are some of the different techniques that have been extensively studied. A comprehensive survey of traditional techniques can be found in ~\cite{borji2015salient}.

Convolutional Neural Networks have been the status quo for vision tasks in recent years, being present in all state-of-the-art methods for Salient Object Detection as well. DSS~\cite{DSS} is currently the best performing method on many benchmark datasets. It is a VGG-based network, where the authors use connections between the deeper and shallower layers, and fuse them in a way similar to HED~\cite{xie2015holistically}. These collective features have the ability to extract necessary high level information, without losing spatial acuity.

In another work by Liu et al.~\cite{DHS}, a coarse-to-fine network is used to improve the features progressively. The authors implement a recurrent structure to gradually increase the spatial precision. Instead of using RNNs for interpolation, we use it to segment the common object(s) from the single image patches. Luo et al.~\cite{NLDF} use a global segmentation branch, and various sub-branches at different levels to extract local features, which are then combined in a separate layer to refine the global prediction. Wang et al.~\cite{wang2017stagewise} use a pyramid pooling scheme to extract multi-scale features before the final prediction layer.

Another work that uses recurrent structure in their architecture is RAN \cite{RAN}, where they used one RNN to predict segments from a local patch, and another RNN conditioned on the first one that proposes the next region in the image to focus on. This work is thematically the most similar to ours. We differ in the approach as we do not use a decoder STN at the end to map the attended region back to input image. We also employ Convolutional GRUs (ConvGRU) instead of vanilla RNNs in our attention module which regresses all regions in one step and incorporates the inverse spatial mapping to input image within the module.

ConvGRU is an extension of a Gated Recurrent Unit~\cite{cho2014properties} which was introduced in ~\cite{ballas2015delving}. A fully convolutional GRU was used for video segmentation by Siam et al.~\cite{siam2017convolutional}. A ConvGRU has been shown to perform well for a spatially structured task with fewer parameters than a traditional GRU.

In most of the other recent methods \cite{amulet,MDF,RFCN}, we find a common practice to combine the lower and higher level features through convolution layers. Due to the lower level features being very noisy, and containing all edges and texture information, these are often combined with the higher level features and passed through convolution filters to suppress the noise. This preserves the stronger edge information, such as the boundary information of the salient object and the weaker noise signals are suppressed.

\begin{figure}[]
\centering
\includegraphics[width=0.9\linewidth]{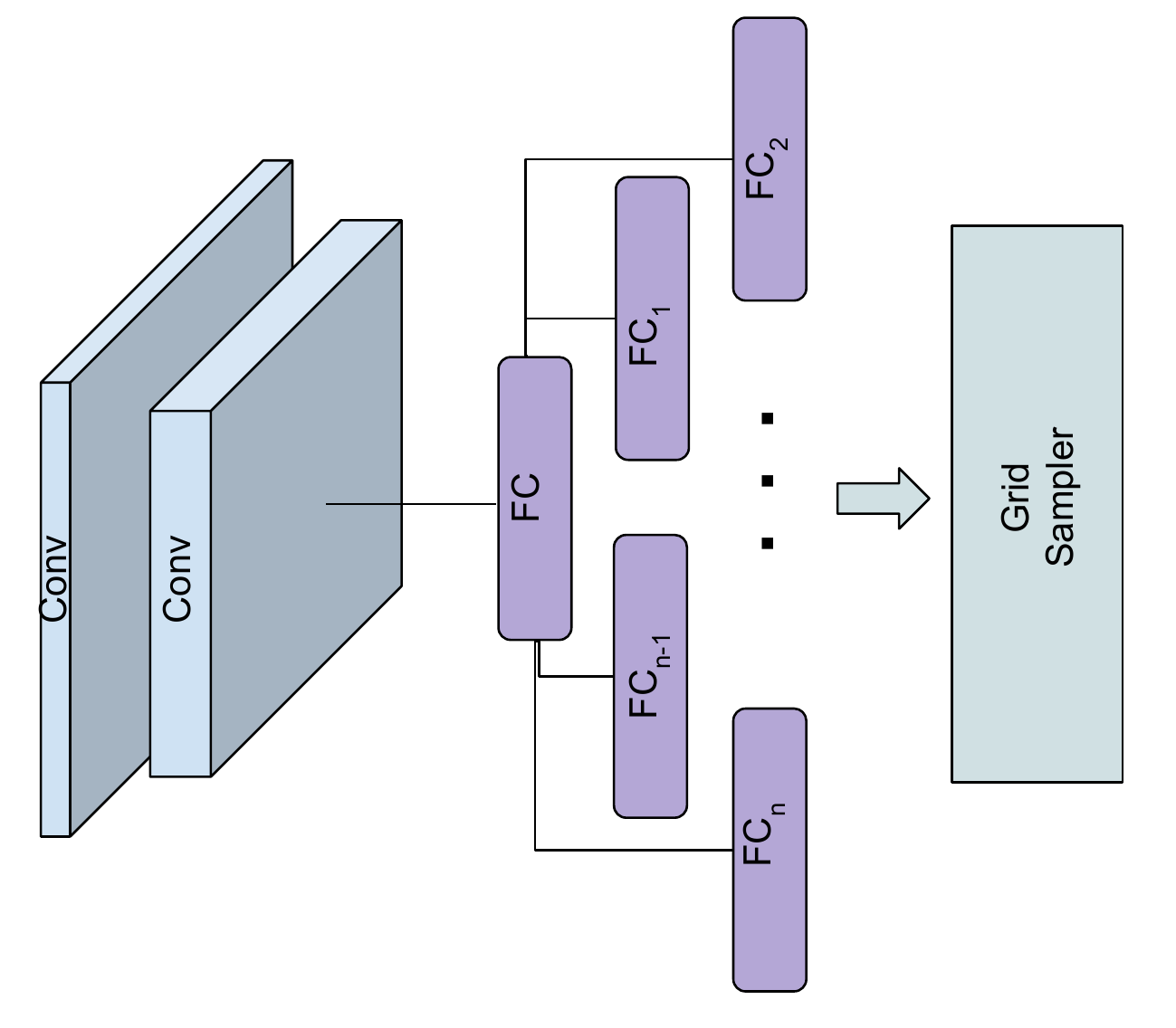}
\caption{Patch Generation Module (PGM).}
\label{fig:pgm}
\end{figure}

\section{Method}

We use a neural network based architecture comprising of CNNs, GRUs and fully connected layers for our method. The architecture is designed to be modular such that each module can be used independently (Figure~\ref{fig:overview}), demonstrating both simplicity and interpretability. We describe the three modules in this section - a Patch Generation Module (PGM), a Saliency Prediction Module (SPM) and a Recurrent Attention Module (RAM).

\subsection{Patch Generation Module (PGM)}
PGM takes an RGB image $I_0$ of dimensions $[H, W, 3]$ as an input and generates $N$ patches per image. It can be considered as a Multi-Spatial Transformer Network (STN)~\cite{jaderberg2015spatial} with a shared localization network barring $N$ fully connected layers (for each patch). The localization network is a small neural network with some convolutional and/or fully connected layers (Figure \ref{fig:pgm}). For our experiments, we use $2$ convolutional layers, each with $64$ kernels of window size $7\times7$ and $5\times5$ respectively. Each $Conv$ layer is followed by a max pooling operation with a stride of $2$. That is followed by a fully connected layer that outputs a $256$-dimensional vector. The output is passed on to $N$ fully connected layers. The $N$ unshared FC layers regress to $4N$ outputs, representing $N$ $[x_1, y_1, x_2, y_2]$ normalized image co-ordinates to crop. 

Unlike a conventional spatial transformer network, we do not regress the parameters for a generic affine transformation. To specifically preserve the spatial appearance of the salient object, we use image crops. Image crops have proven to be good `perturbations' for an iterative refinement task~\cite{fu2017look}. Even though affine perturbations add more degrees of freedom, they are well suited for an image-level classification task where preserving the spatial structure is less important. We further validated this experimentally and found the supervisory signal to be too weak to train the complete STN.

Spatial context is another important cue for good detection. Without a spatial context, the saliency features generated might not be optimal. Thus, we explicitly enforce $x_2 - x_1 \geq \epsilon$ and $y_2 - y_1 \geq \epsilon$, where $\epsilon$ is the percentage of image to crop. This also gives a good initialization to the model due to the lack of direct supervision on the crop parameters. We empirically choose $\epsilon=0.6$ for our experiments as it is a good trade-off between the amount of zoom we want for a patch and the amount of overlap that could provide a good association of spatial context for our recurrent module.

The generated $4N$ crop parameters are used to crop $I_0$ and resized to fixed dimensions $[H, W, 3]$ through a differentiable grid sampling layer. The $N$ generated patches are passed along with $I_0$, the full image, as a batch of ($N+1$) images ($[I_0, I_1, \dots, I_N]$) to the next module.

\begin{figure}[]
\centering
\includegraphics[width=0.9\linewidth]{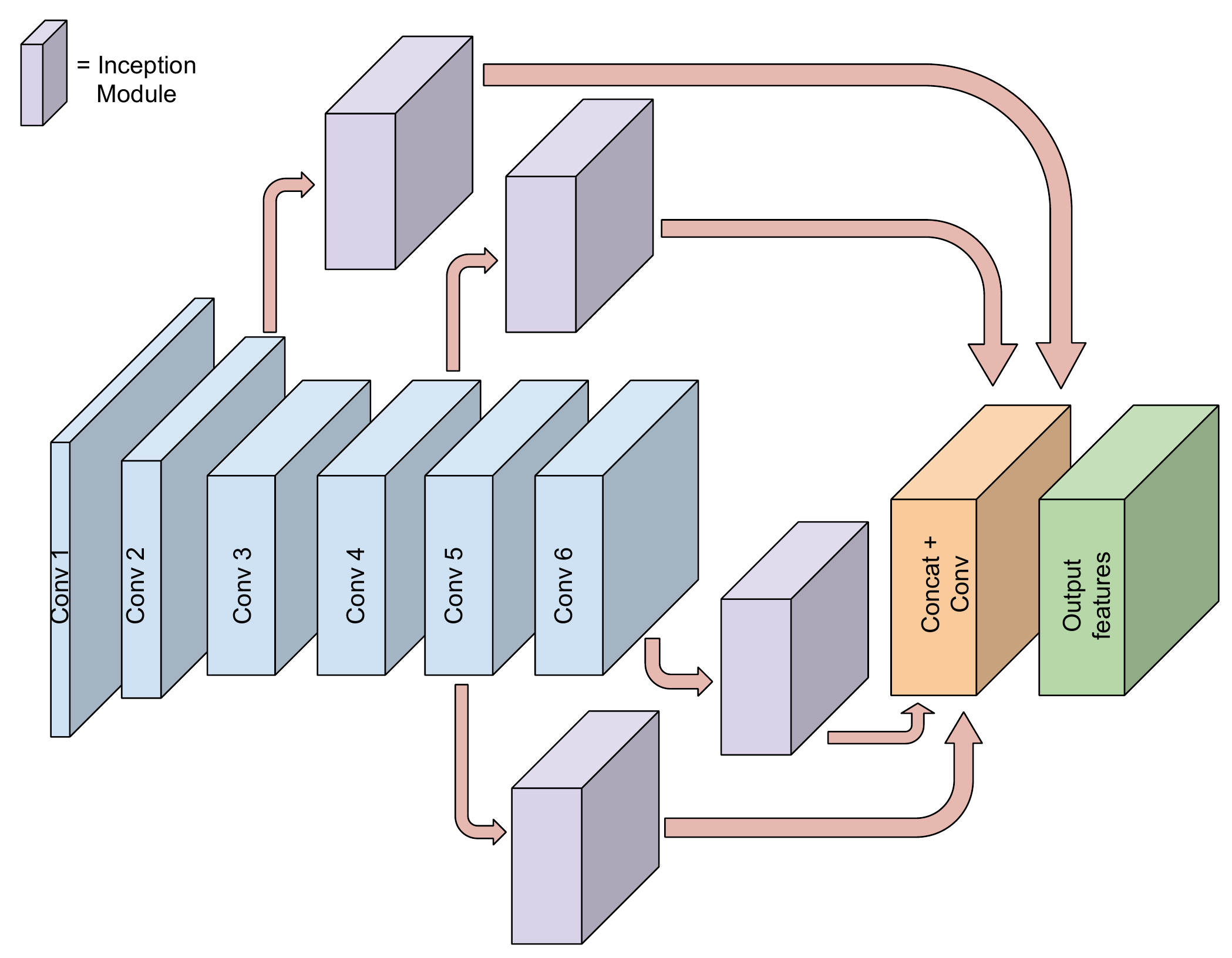}
\caption{Saliency Prediction Module}
\label{fig:spm}
\end{figure}

\begin{figure}[]
\centering
\includegraphics[width=\linewidth]{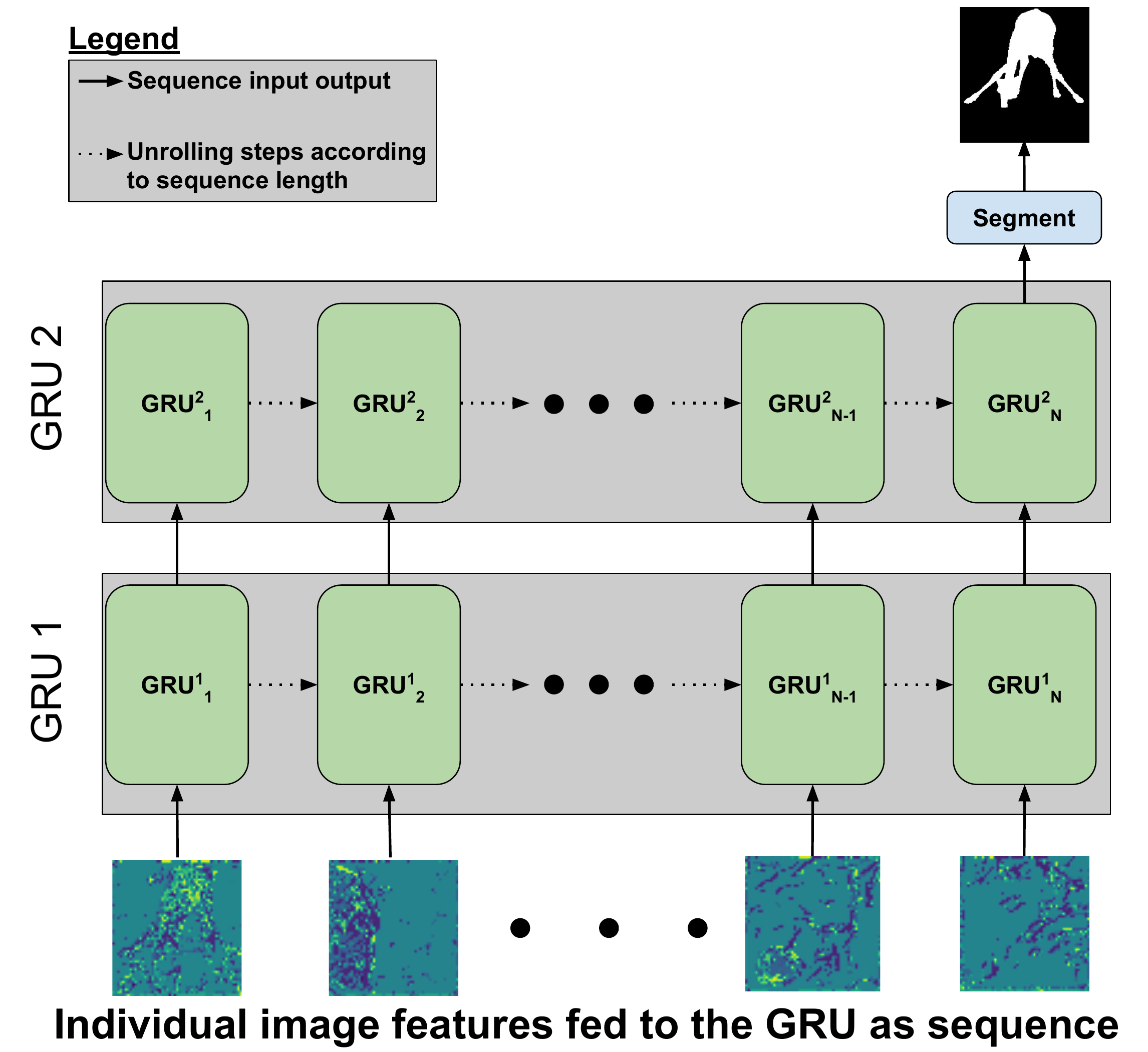}
\caption{Recurrent Attention Module (RAM). The ConvGRUs in RAM have been unrolled for visualization.}
\label{fig:modules}
\end{figure}

\subsection{Saliency Prediction Module (SPM)}
\label{section:SPM}
SPM is the primary saliency feature extractor in our architecture. We base our SPM network on the segmentation branch of Saliency Unified~\cite{SU}, which is a modified VGG-16~\cite{simonyan2014very} network that achieved impressive results in saliency prediction task. The convolutional part of the original VGG-16 network makes the input image $1/32$ of the original size, which makes the task of spatially localizing the objects imprecise. To overcome this, the final two convolution layers of VGG-16 ($Conv4$ and $Conv5$) are replaced with dilated convolutional layers~\cite{yu2015multi}. Dilated convolutions span a bigger field of view while effectively preserving the spatial resolution and maintaining the same number of parameters. Unlike~\cite{SU}, we introduce only one additional convolution layer abstraction after the $Conv5$ layer. Inception modules are used to fuse features from different layers at multiple scales. The network finally outputs a $512$-dimensional feature blob at $1/8$ resolution of the original image.

A batch of ($N+1$) $512$-dimensional feature maps ($[F_0, F_1, \dots, F_N]$), corresponding to the ($N+1$) image patches is generated and then passed on to the next module.

\subsection{Recurrent Attention Module (RAM)}
\label{section:RAM}
The task of RAM is to aggregate the learned bag of features from the previous module in a semantically coherent manner and improve the final segmentation map. RAM is implemented as two Convolutional GRUs~\cite{convgru2015} in an encoder-decoder style. A ConvGRU has fewer parameters than a traditional GRU and performs well on spatially structured data.

A ConvGRU is comprised of convolutional layers as opposed to fully connected layers in a traditional GRU. The function of hidden units is identical to the normal GRU and can be represented as: 
\begin{equation}
z_t = \sigma_g(W_z * x_t + U_z * h_{t-1} + b_z),
\end{equation}
\begin{equation}
r_t = \sigma_g(W_r * x_t + U_r * h_{t-1} + b_r),
\end{equation}
\begin{equation}
h_t = (1-z_t)\odot h_{t-1} + z_t \odot \sigma_h(W_h * x_t + U_h*(r_t \odot h_{t-1}) + b_h),
\end{equation}
where $*$ represents the convolution operation, $z_t$ is the update gate and $r_t$ is the reset gate. Unlike LSTM~\cite{hochreiter1997long}, GRU has only two gates and no internal state. $W,U,b$ are the training parameters of the GRU, and $x_t$ and $h_t$ are input and output activation blobs respectively.

Figure~\ref{fig:modules} depicts a simplistic view and workflow of RAM. We use convolution filters of size $5\times5$ for the GRUs, slightly larger than the size of filters in SPM ($3\times3$). This helps the ConvGRUs to learn a distinction between background and foreground regions at $1/8$ scale as a $5\times5$ filter will mostly see the background and foreground together. While the architecture is not exactly an encoder-decoder, $2$ GRUs are used to make the learned representations co-dependent similar to what an encoder-decoder setup does. Since the $2^{nd}$ GRU directly outputs the saliency maps, it acts like a decoder. Furthermore, this specific architecture allows us to avoid the need for an inverse spatial transformer as the one used in ~\cite{RAN}. We finish the network with a $1\times1$ convolution filter to scale the output of the decoder GRU to get the final pixel-wise predictions.

We order the incoming ($N+1$) feature maps such that the input image feature map is fed first into the module (Figures \ref{fig:overview}, \ref{fig:modules}). This is done to ensure that RAM gets to learn the complete spatial context first. The task is then reduced to learning spatial associations of incoming feature maps and predict a saliency map for each recurrent step. Every feature map $F_k$, where $k \in [0, N]$ is iteratively fed into RAM where the decoder GRU outputs a saliency map $Pred_k$. We enforce the iterative improvement criterion by weighing the loss for $Pred_{k+1}$ more than the loss for $Pred_{k}$. Refer to next section (Sec \ref{section:setup}) for more details.

\subsection{Implementation}
\label{section:setup}
The initial layers of Saliency Prediction Module are initialized using pre-trained ImageNet weights of VGG-16. Rest of the its layers are initialized with Xavier initialization scheme~\cite{glorot2010understanding}. PGM and RAM are also both initialized using the same scheme. The training is carried out in a step-wise manner. We first train SPM for object segmentation using the training datasets. Since the proposed SPM (Figure~\ref{fig:spm}) only outputs a feature blob, we add $3$ convolutional layers after it that decode and predict a saliency map. We use this prediction to fully train SPM. We minimize the loss function - 
\begin{eqnarray}
L = \lambda_1Loss_{CE} + \lambda_2Loss_{IOU},
\end{eqnarray}
where $Loss_{CE}$  is the standard sigmoid cross-entropy loss and $Loss_{IOU}$ is an IOU-based loss described in~\cite{rahman2016optimizing}. $\lambda_1$ and $\lambda_2$ are kept as $1.0$. We use a batch size of 10 and optimize it using Adam~\cite{kingma2014adam} with a learning rate of $1e^{-5}$. We decrease the learning rate step-wise based on the validation performance. We train it for $10$ epochs.

We then take out the added convolutional layers from SPM and plug the $512$-dimensional feature blob to RAM. We also plug in PGM by placing it before SPM so that it outputs ($N+1$) patches for input image $I_0$. These are then passed on to SPM as a batch. We use a batch size of $1$ in this complete setup. We freeze SPM layers during the training. We optimize the whole network on an exponentially weighted loss on RAM's outputs-

\begin{eqnarray}
L = \frac{1}{k^{N+1}}\sum_{i=0}^{N} k^{i+1}Loss_{CE_i},
\end{eqnarray}
where $i \in [0, N]$ and value of $k$ is chosen to be $2$. $Loss_{CE_i}$ is the sigmoid cross-entropy loss between $Pred_i$ and ground truth label. The described loss gives more weight to every ${i+1}^{th}$ prediction compared to $i^{th}$ as described in Section \ref{section:RAM}. Adam optimizer is used with a learning rate of $1e^{-4}$ for PGM and $5e^{-6}$ for the RAM. This setup is trained for 10 epochs. We further fine-tune the complete network end-to-end for 5 epochs. We use $N = 4$ for our experiments.

During testing, we adopt the same approach as our training mechanism. For a single image, we get ($N+1$) predictions from RAM. We resize the predictions to the original image size using bilinear interpolation. We use $Pred_5$ (prediction after seeing all patches) for our final performance evaluations. 

\section{Experiments}

\subsection{Datasets}
We use the Pascal VOC-2012~\cite{pascal-voc-2012} and MSRA10K~\cite{msra} datasets for training our model and ECSSD~\cite{ecssd}, HKU-IS~\cite{hku}, DUT-OMRON~\cite{omron} and PASCAL-S~\cite{li2014secrets} for evaluation. 300 random images from the training dataset are used for validation.
\newline
\textbf{PASCAL-VOC 2012.} Pascal-VOC dataset has semantic segmentations of 20 object classes. We convert these into binary segmentation maps and use for our task. This dataset has scenes containing complex background and multiple salient objects in the scene.
\newline
\textbf{MSRA10K.} This dataset contains 10000 image-label pairs of salient objects in varied scenes.
\newline
\textbf{DUT-OMRON.} This dataset consists of 5168 high quality images featuring one or more salient objects and relatively complex background.
\newline
\textbf{ECSSD.} ECSSD  contains 1000 images of natural scenes, often comprising semantically meaningful and complex structures to segment.
\newline
\textbf{HKU-IS.} This dataset contains 4447 images with high-quality object annotations. Many images include multiple disconnected objects or objects touching the image boundary.
\newline
\textbf{PASCAL-S.} PASCAL-S is the testing subset of 850 images from the PASCAL VOC dataset.

\subsection{Evaluation metrics}
\label{section:eval_metrics}
One of the evaluation metrics for the image dataset is Mean Absolute Error or MAE. The MAE computes the average pixel percent error. It is computed as:

\begin{eqnarray}
{MAE = \frac{1}{M \times N}  \sum_{i,j} |G(x_{ij})-P(x_{ij})| }
\end{eqnarray}
where $x_{ij}$ is the input image of width and height M and N, $G(\cdot)$ and $P(\cdot)$ are the ground truth mask and predicted mask of the input image respectively.

The other metric is $F_\beta$ score, which is a weighted ratio of the Recall and Precisions. The recall is computed as $TP/(TP + FN)$ and the precision is computed as $TP/(TP + FP)$. Here $TP$, $FP$ and $FN$ hold their usual meanings of True Positive, False Positive and False Negative predictions respectively. Then $F_\beta$ score can then be computed as:
\begin{eqnarray}
F_\beta = \frac{(1+\beta^2)\times Precision \times Recall}{\beta^2 \times Precision + Recall},
\end{eqnarray}
where $\beta^2 = 0.3$ to weigh precision more than recall rate~\cite{yang2013saliency}. This is done in accordance with the number of negative examples (non-salient pixels) typically being much bigger than the number of positive examples (salient object pixels) while  evaluating SOD  models. Hence, $F_\beta$ score is a good indicator of an algorithm's detection performance~\cite{borji2015salient}. We compare against the maximum $F_\beta$ scores of all other approaches.

\begin{table*}[h!]
\begin{center}
\begin{tabular}{|c|cc|cc|cc|cc|}
\hline
\multirow{2}{*}{Methods} & \multicolumn{2}{c|}{DUT-OMRON} &
\multicolumn{2}{c|}{HKU-IS} &
\multicolumn{2}{c|}{ECSSD} &
\multicolumn{2}{c|}{PASCAL-S}\\
\cline{2-9}
& MAE & $max. F_\beta$ & MAE & $max. F_\beta$ & MAE & $max. F_\beta$ & MAE & $max. F_\beta$\\
\hline \hline

BSCA \cite{qin2015saliency} & - & - & 0.175 & 0.719 & 0.182 & 0.758 & 0.223 & 0.667 \\  

DRFI \cite{jiang2013salient} & - & - & 0.145 & 0.777 & 0.164 & 0.786 & 0.207 & 0.698 \\

RFCN \cite{RFCN} & - & - & 0.079 & 0.892 & 0.107 & 0.890 & 0.118 & 0.837 \\

DHS \cite{DHS} & - & - & 0.053 & 0.890 & 0.059 & 0.907 & 0.094 & 0.829 \\

DCL \cite{DCL} & 0.084 & 0.733 & 0.054 & 0.892 & - & - & 0.113 & 0.815 \\

UCF \cite{zhang2017learning} & 0.080 & 0.726 & 0.074 & 0.886 & 0.078 & 0.911 & 0.126 & 0.828 \\

Amulet \cite{amulet} & 0.074 & \textbf{0.741} & 0.052 & 0.895 & 0.059 & 0.915 & 0.098 & \textbf{0.837} \\

SRM \cite{wang2017stagewise} & 0.069 & 0.707 & \textbf{0.046} & 0.874 & \textbf{0.056} & 0.892 & - & - \\

NLDF \cite{NLDF} & 0.085 & 0.724 & 0.060 & 0.874 & 0.075 & 0.886 & 0.108 & 0.804 \\

DSS* \cite{DSS} & \textbf{0.068} & 0.736 & \textbf{0.039} & \textbf{0.913} & \textbf{0.052} & \textbf{0.916} & \textbf{0.080} & 0.830 \\

Ours & \textbf{0.066} & \textbf{0.751} & 0.054 & \textbf{0.915} & 0.063 & \textbf{0.921} & \textbf{0.083} & \textbf{0.846} \\

\hline
\end{tabular}
\end{center}
\caption{Quantitative comparison with other state-of-the-art methods on various datasets. Top two results are in \textbf{bold} numbers.}
\label{tab:results}
\end{table*}

\subsection{Comparison with existing approaches}
In Table \ref{tab:results}, we compare the quantitative performance of various state-of-the-art methods with ours based on the aforementioned evaluation criteria. We compare against DSS~\cite{DSS}, DCL~\cite{DCL}, DHS~\cite{DHS}, Amulet~\cite{amulet}, SRM~\cite{wang2017stagewise}, UCF~\cite{zhang2017learning}, RFCN~\cite{RFCN} and two non-deep methods - DRFI~\cite{jiang2013salient} and BSCA~\cite{qin2015saliency}.

Our method consistently gets top $F_{\beta}$ scores, implying a greater precision in the predicted map. The high precision showcases its effectiveness on suppressing false positives in cluttered backgrounds and partly salient objects. Metrics of other methods have either been reported by the respective authors or have been computed by us using available predictions/weights. For a fair comparison, we use the scores obtained without post-processing for all methods. 

\blfootnote{*DSS also employs a CRF post-processing step.}

\begin{figure*}[]
\centering
\includegraphics[width=1.0\linewidth]{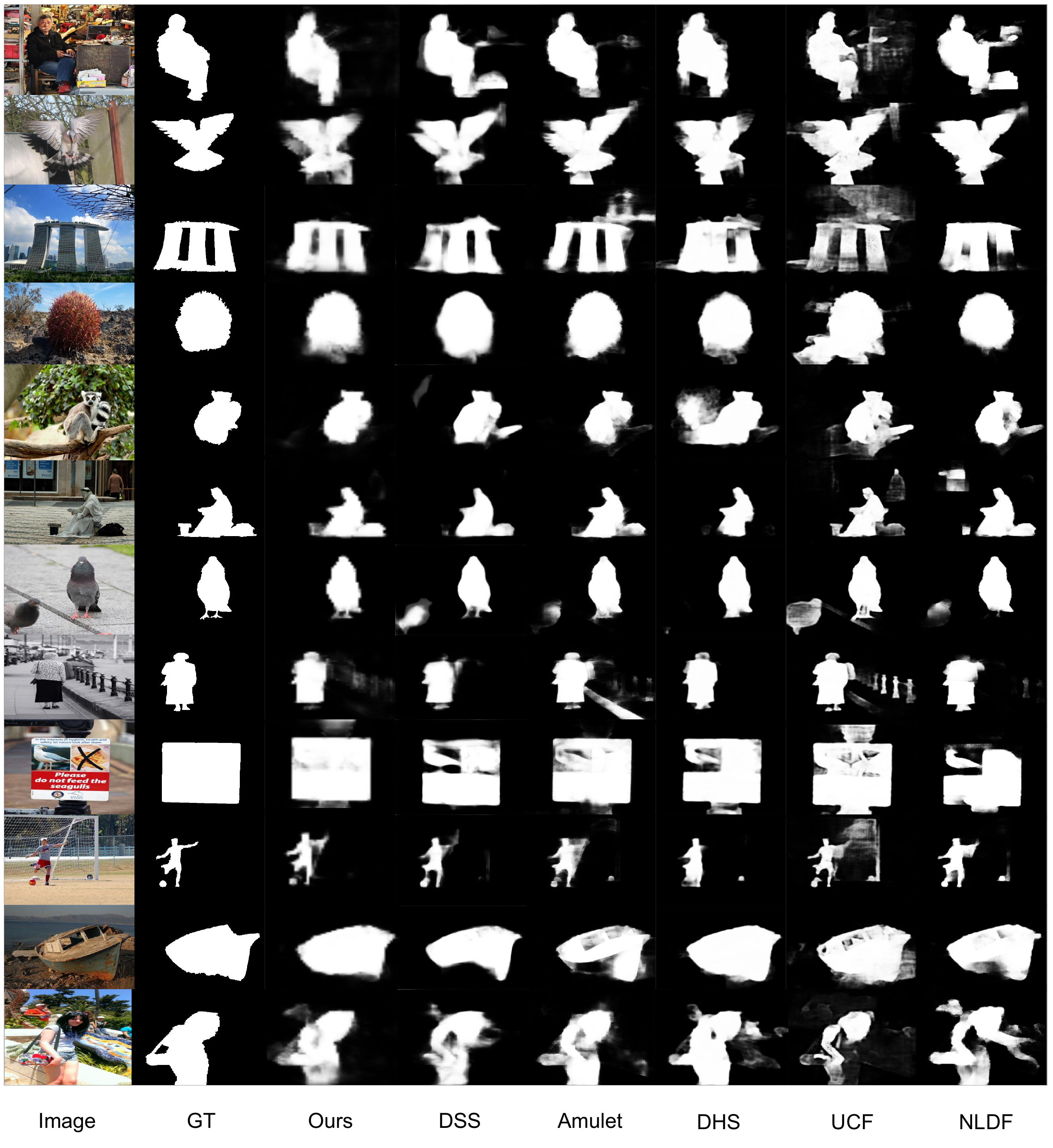}
\caption{Qualitative comparison with various state-of-the-art approaches on some challenging images from ECSSD~\cite{ecssd}. Most of the images where we perform better are the ones where global spatial context is important to distinguish between foreground and background.}
\label{fig:qual}
\end{figure*}

In Figure \ref{fig:qual}, we compare the qualitative results of the aforementioned methods with ours. We show results for a set of images with varying difficulties:
\newline
\textbf{Cluttered background.} Row $1$ contains a textured background, making algorithms prone to background noise.
\newline
\textbf{Shadows in background.} Rows $2$ and $4$ include images with object shadows. While every method performs well on Row $4$, our method is able to suppress much of the `shadow saliency' in background of Row $2$ that is easily thresholded during inference.
\newline
\textbf{Low contrast.} Row $7$ contains an image with low contrast between object and background. We are able to segment better with fewer false positives than others.
\newline
\textbf{Multiple Objects.} Rows 5, 6, 7 and 10 contain multiple foreground objects. 6 and 10 contains multiple salient objects whereas 5 and 7 have only a single salient object. Our algorithm performs very well in these scenarios.
\newline
\textbf{Complex foreground.} Row 12 contains a complex salient object where most other algorithms create holes in the prediction. Our method is able to better understand the regional and global context.
\newline
\textbf{Object within an object.} Row 9 contains an interesting image which contains an image of a bird (with sharp contrast) within a poster (salient object). Our method is the only method that does not fail by trying to segregating these two objects.

\subsection{Method Analysis}
\label{section:ablation}
We analyze our network's performance by evaluating component-wise and step-wise results. The results shed light on our design choices and incremental gains. The evaluation metrics have been described in Section \ref{section:eval_metrics}.


\begin{figure*}[]
\centering
\includegraphics[width=0.8\linewidth]{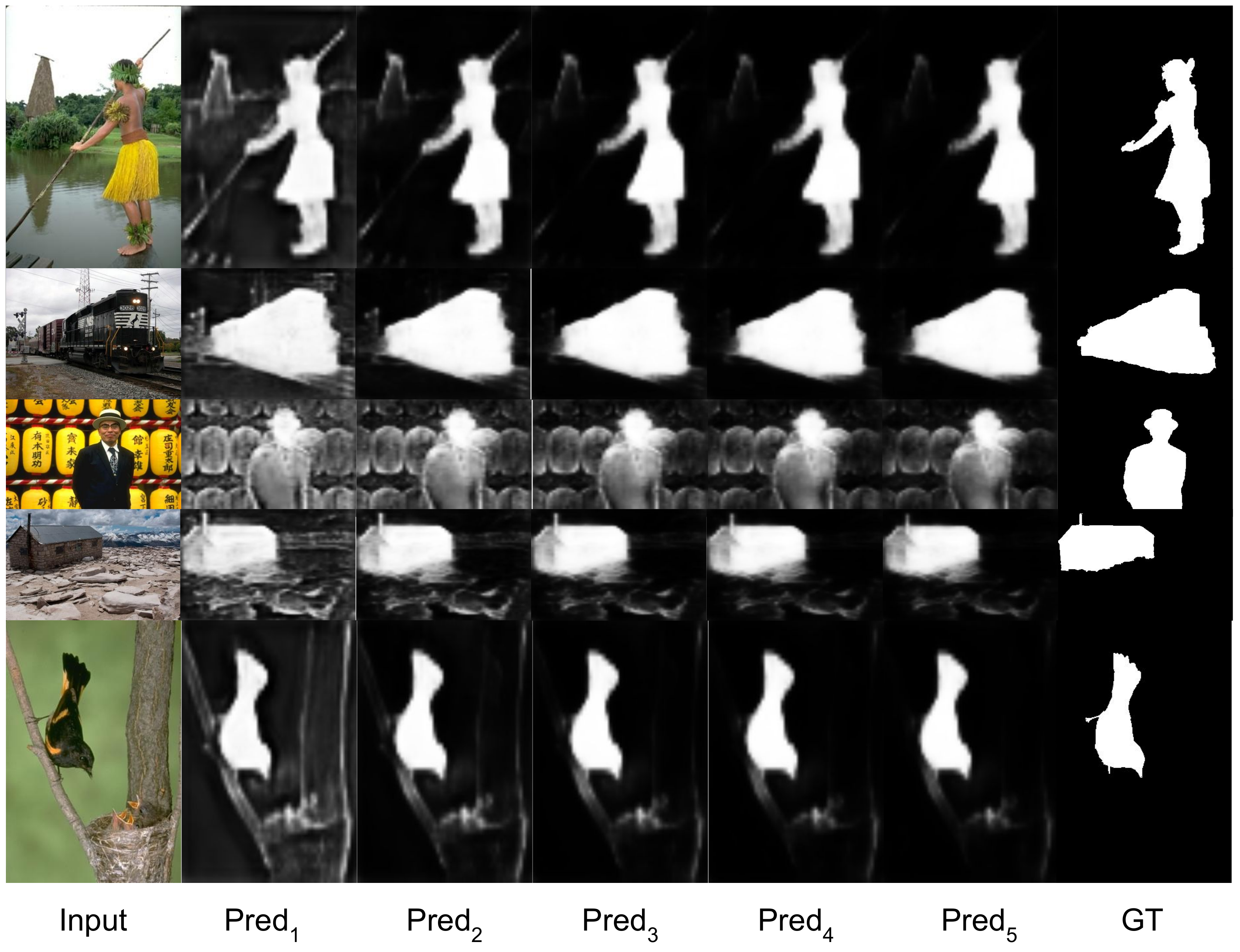}
\caption{Recurrent step-wise qualitative performance analysis. We observe that $Pred_1$ captures a lot of `pseudo' salient objects. As we go from left to right, we see a clear reduction in number of false positives that arise from background.}
\label{fig:qual-ablation}
\end{figure*}

\setlength{\tabcolsep}{4pt}
\begin{table}[t]
\begin{center}
\begin{tabular}{ccc}
\hline\noalign{\smallskip}
Module & MAE $(\downarrow)$ & $max. F_\beta$ $(\uparrow)$\\
\noalign{\smallskip}
\hline
\noalign{\smallskip}
SPM & 0.080 & 0.870 \\
SPM + RAM (Epoch 2) & 0.0692 & 0.9193\\
PGM + SPM + RAM (Epoch 2) & 0.0662 & 0.9196\\
SPM + RAM (Epoch 5) & 0.0661 & 0.9205\\
PGM + SPM + RAM (Epoch 5) & 0.0623 & 0.9204\\
\hline
\end{tabular}
\end{center}
\caption{Incremental performance gains for different modules on ECSSD}
\label{tab:mod_ablation}
\end{table}
\setlength{\tabcolsep}{1.4pt}

\setlength{\tabcolsep}{4pt}
\begin{table}[t]
\begin{center}
\begin{tabular}{ccc}
\hline\noalign{\smallskip}
$Pred_k$ (Epoch 2) & MAE $(\downarrow)$ & $F_\beta$ $(\uparrow)$ \\
\noalign{\smallskip}
\hline
\noalign{\smallskip}
$k=1$ & 0.0692 & 0.9193\\
$k=2$ & 0.0687 & 0.9196\\
$k=3$ & 0.0666 & 0.9197\\
$k=4$ & 0.0669 & 0.9197\\
$k=5$ & 0.0662 & 0.9196\\
\hline
\end{tabular}
\begin{tabular}{ccc}
\hline\noalign{\smallskip}
$Pred_k$ (Epoch 5) & MAE $(\downarrow)$ & $F_\beta$ $(\uparrow)$\\
\noalign{\smallskip}
\hline
\noalign{\smallskip}
$k=1$ & 0.0661 & 0.9205\\
$k=2$ & 0.0637 & 0.9210\\
$k=3$ & 0.0629 & 0.9208\\
$k=4$ & 0.0625 & 0.9206\\
$k=5$ & 0.0623 & 0.9204\\
\hline
\end{tabular}
\end{center}
\caption{Quantitative performance comparison of $N$ predictions from the Recurrent Attention Module on ECSSD}
\label{tab:out_ablation}
\end{table}
\setlength{\tabcolsep}{1.4pt}


To better quantify the role of every module in our architecture, we do a component-wise performance analysis on ECSSD dataset (Table \ref{tab:mod_ablation}). We first compute the results using just SPM with added convolutional layers as described in Section \ref{section:setup}. We can easily evaluate its performance independently since it is trained first. We then plug in RAM to SPM's $512$-dimensional features and do an inference on trained SPM and RAM by only evaluating on $Pred_0$. We see an immediate performance boost with this setting. While this could just be attributed to increase in model complexity, we argue that the initial setup with SPM + 3 layers also has similar complexity. This observation shows that RAM not only predicts a better output for every $Pred_{k+1}$($k \in [0, N-1]$), but also improves $Pred_i$($i \in [0, k]$) in the process. We do a final evaluation by allowing a forward pass through all three modules.

In a recurrent network, we should ideally see performance improvements for every iterative step. To verify this, we evaluate the predicted saliency maps computed at every $k^{th}$ step and compare the results in Table \ref{tab:out_ablation}. We evaluate results after $2^{nd}$ and $5^{th}$(final) epoch. For both the readings, we observe that the $F_\beta$ scores do not vary much across the ($N+1$) predictions. Higher $F_\beta$ does not imply a lower MAE~\cite{borji2015salient}. Often, a decrease in MAE also leads to a decrease in $F_\beta$. Thus, an important observation is the gradual decrease in MAE as we increase $k$. A decrease in MAE while maintaining the $F_\beta$ scores affirms that RAM reduces the false positives incrementally without losing precision.

Qualitatively, we observed that visible differences in saliency maps are more noticeable during the initial epochs. Hence, we show the ($N+1$) predicted maps after the $1^{st}$ epoch in Figure~\ref{fig:qual-ablation}. We can clearly notice the suppression of false positives in background for every subsequent prediction. 

\section{Conclusion}
We present an intuitive, scalable and effective approach for detecting salient objects in a scene. Our approach is modular, resulting in interpretable results. We propose a Patch Generation Module, a Saliency Prediction Module and a Recurrent Attention Module that work in tandem to improve overall object segmentation by generating image patches, their corresponding feature maps and effectively aggregating them. Through our quantitative and qualitative performance on benchmark datasets, we show the importance of region-wise attention in saliency prediction. An easy and important extension to our work could be a dynamic improvement of predictions based on the number of allowed patches. This can reduce the inference time significantly for an accuracy trade-off. In future, we would also like to test our method's effectiveness on the task of video object segmentation in an unsupervised setting.

\clearpage

{\small
\bibliographystyle{ieee_fullname}
\bibliography{egbib}
}

\end{document}